\documentclass[]{interact}

\usepackage{epstopdf}
\usepackage{subfigure}

\usepackage{natbib}
\bibpunct[, ]{(}{)}{,}{a}{}{,}
\renewcommand\bibfont{\fontsize{10}{12}\selectfont}

\usepackage{lineno,xcolor}
\usepackage{amsmath,amssymb}
\usepackage{algorithm}
\usepackage{algpseudocode}
\usepackage{booktabs}
\usepackage{multirow}
\usepackage{subcaption}
\usepackage{tabularx}
\usepackage{url}
\usepackage{hyperref}

\theoremstyle{plain}

\theoremstyle{definition}

\theoremstyle{remark}

\begin{document}

\articletype{ARTICLE}

\title{SMOL-MapSeg: Show Me One Label as prompt}

\author{
\name{Yunshuang Yuan\textsuperscript{a}\thanks{Corresponding author: Yunshuang Yuan. Email: yunshuang.yuan@ikg.uni-hannover.de}, Frank Thiemann\textsuperscript{a}, Thorsten Dahms\textsuperscript{b}, Monika Sester\textsuperscript{a}}
\affil{\textsuperscript{a}Leibniz University Hannover, Hanover, Germany; \textsuperscript{b}The German Federal Agency for Cartography and Geodesy (BKG)}
}

\maketitle

\begin{abstract}
Historical maps offer valuable insights into Earth’s surface changes but pose challenges for modern segmentation models due to inconsistent visual styles and symbols. While deep learning models like UNet and pre-trained foundation models perform well in domains such as autonomous driving and medical imaging, they struggle with the variability of historical maps, where similar concepts appear in diverse forms. To overcome this, we propose On-Need Declarative (OND) knowledge-based prompting, a method that provides explicit image-label pair prompts to guide models in linking visual patterns with semantic concepts. This allows users to define and segment target concepts on demand, enabling flexible, concept-aware segmentation. Our approach replaces the prompt encoder of the Segment Anything Model (SAM) with the OND prompting mechanism and fine-tunes it on historical maps, creating SMOL-MapSeg (Show Me One Label\footnote{“One Label imaged” refers to the single image-label pair used as a prompt during inference and is \textbf{not} related to one-shot fine-tuning.}). Unlike existing SAM-based fine-tuning methods that are class-agnostic or limited to fixed classes, SMOL-MapSeg supports class-aware segmentation across arbitrary datasets. Experiments show that SMOL-MapSeg accurately segments user-defined classes, significantly outperforming baseline models. Moreover, it demonstrates strong generalization, even with minimal training data, highlighting its potential for scalable and adaptable historical map analysis.
\end{abstract}


\section{Introduction}

\begin{figure}
  \includegraphics[width=\textwidth]{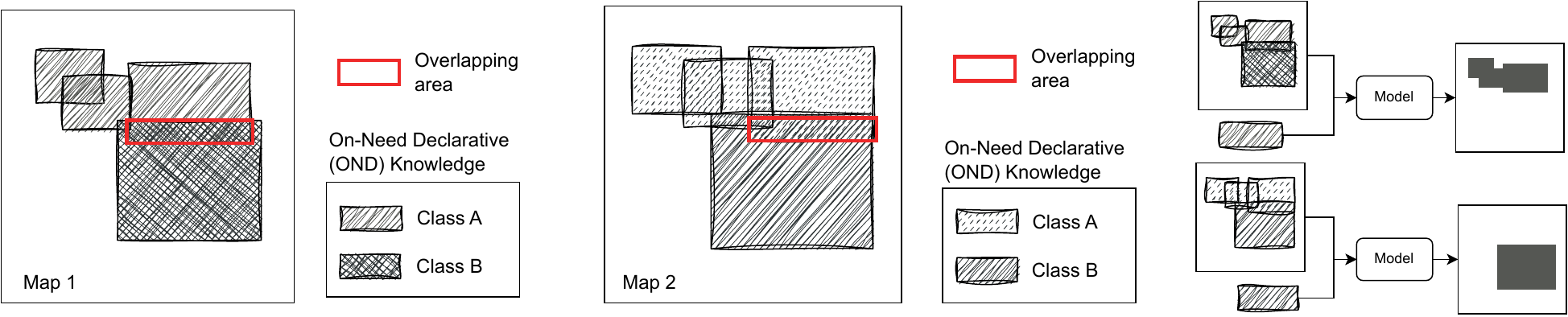}
  \caption{The concept of segmenting images based on a newly provided labeling example (OND Knowledge). In historical maps, different maps may use different visual patterns to represent the same class (e.g., Map 1 and Map 2 using different patterns for Class A), or conversely, the same pattern to represent different classes (e.g., Class A in Map 1 and Class B in Map 2). Overlapping patterns may introduce additional complexity. This inconsistency can confuse a conventional semantic segmentation model. By prompting the model with OND knowledge—which specifies “what something is” in the current context—this ambiguity can be effectively resolved (see illustration on the right).}
  \label{fig:teaser}
\end{figure}

\begin{figure*}[ht]
    \centering
    \includegraphics[width=\linewidth]{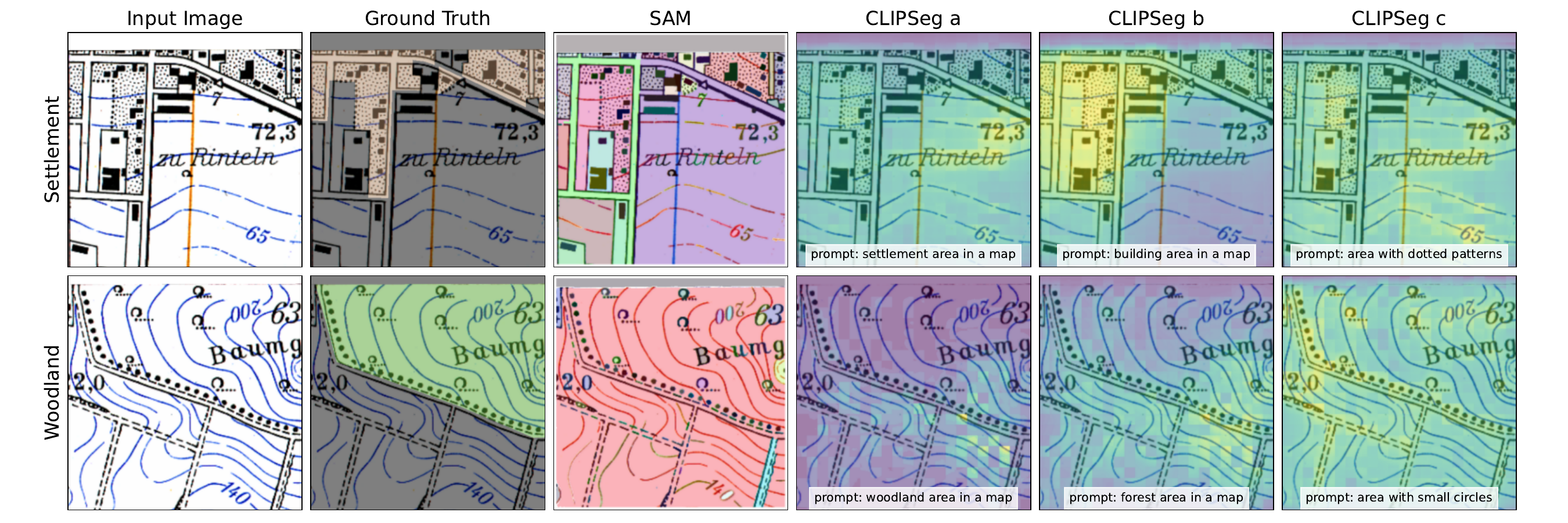}
    \caption{Examples of segmentation results of historical maps by SAM and CLIPSeg.}
    \label{fig:sam_clip}
\end{figure*}

Historical maps offer valuable insights into past conditions of the Earth's surface, including changes over time driven by both natural processes and human activities. These maps are typically preserved as scanned raster images, which require further processing to extract semantic information. This extracted data can then be conveniently queried for various applications, such as statistical analyses of changes in the Earth's surface. To automate the information extraction process, recent studies have utilized deep learning techniques—particularly Convolutional Neural Networks (CNNs)—for semantic segmentation of historical maps~\citep{Lukas_semseg,maxwell_semantic_2020,jiao_extracting_2020,guo_semantic_2018,petitpierre_generic_2021}. However, these models typically require large amounts of labeled training data, with ground-truth annotations that are costly and time-consuming to produce manually. To reduce the need for manual annotation, \cite{uhl_automated_2020}, \cite{yuan2025semseg} and \cite{wu_domain_2023} explored spatial co-occurrence of geo-located map features across time as a weak supervisory signal for segmenting historical maps. Although effective, these models are trained on a specific map collection with similar figurative features and can hardly be generalized for large scale utilization.

To train generalized models for multi-domain semantic segmentation tasks, foundation models such as SAM~\citep{sam} and CLIPSeg~\citep{clipseg} have been developed. These models demonstrate impressive performance in various everyday scenarios; however, they perform poorly on historical maps. Figure~\ref{fig:sam_clip} illustrates segmentation results from these two models. The task is to identify settlement and woodland areas, marked in the ground-truth masks (second column) with orange and green, respectively. Both SAM and CLIPSeg fail to produce accurate masks that align with the ground truth due to the significant domain gap between the abstract visual patterns of (historical) maps, which are inherently symbolic, and the conceptual knowledge learned from modern imagery.

SAM and CLIPSeg are primarily trained on photographs from daily life, where semantic classes often share consistent colors, shapes, or textures. These models rely on common-sense visual concepts broadly understood by humans. However, such general prior knowledge is insufficient for accurately interpreting historical maps, which, unlike pictures, do not look realistic, and where the classes are described by symbols that are often not immediately understood, and are also different in maps of different times or/and countries. Without referencing to the map legends, it can be extremely challenging to assign specific patterns to semantic classes.

As shown in Figure~\ref{fig:teaser}, different historical maps may use the same pattern to denote different classes (e.g., Class A in Map 1 vs. Class B in Map 2) or different patterns for the same class (e.g., Class A in both maps). Additionally, patterns may overlap or blend, as seen in the red box of Figure~\ref{fig:teaser}, complicating semantic segmentation. In such cases, on-the-fly interpretation is required to correctly delineate target areas. For instance, in the first row of Figure~\ref{fig:sam_clip}, we define settlements as areas with dotted patterns and black building symbols, while excluding regions that only feature building symbols, as these are considered to be industrial zones. Such definitions are highly variable and often context-dependent, reflecting task-specific objectives in different historical map–analysis scenarios.

To address this, we propose a knowledge taxonomy comprising \textit{prior knowledge},  \textit{on-need declarative (OND) knowledge} and \textit{procedural knowledge}.
Prior knowledge refers to generalized, task-agnostic patterns learned from large-scale data. OND knowledge represents task-specific or conditional definitions—knowledge that tells the model "what something is" in a particular context. Procedural knowledge denotes the model’s ability to execute a task—how to perform it—by leveraging both prior knowledge and the provided OND knowledge. As shown on the right side of Figure~\ref{fig:teaser}, given the new input image and the OND knowledge, the model will be able to capture the prior and procedural knowledge after the training and solve the task accordingly.

In the context of large language models (LLMs), OND knowledge can be regarded as prompt to elicit a specific behavior from the model. Similarly, models like SAM utilize prompts that provide locational cues to identify target regions for segmentation. However, in the case of historical maps, we propose incorporating OND knowledge through example-based prompting. Specifically, we provide a labeled source image that includes the desired pattern as an explicit example. The model is then tasked with identifying and locating the same pattern in a separate target image. This form of OND prompting encodes the task-specific definition of the target class and helps to bridge the domain gap by explicitly grounding the model’s understanding in concrete visual references.


\begin{itemize}
\item We introduce an On-Need Declarative (OND) knowledge-based prompting method that steers foundation models with class-specific image-label pairs, enabling, for the first time, class-aware segmentation across arbitrary classes and datasets.

\item We develop and train SMOL (Show-Me-One-Label)-MapSeg, a modified version of SAM, for generalized semantic segmentation on historical maps. SMOL-MapSeg effectively segments target classes defined by OND prompts and achieves higher average IoU than the baseline models.

\item We adapt SMOL-MapSeg to previously unseen classes through few-shot fine-tuning, demonstrating its strong generalization capability.
\end{itemize}

\section{Related Work}
\subsection{Semantic segmentation}
Semantic segmentation is the task of assigning a class label to each pixel in an image. It yields dense predictions with per-pixel precision, providing fine-grained understanding of the given image. The seminal work by Long et al. introduced the Fully Convolutional Network~\citep{long2015fcn}, which marked the beginning of end-to-end deep learning methods for semantic segmentation. It uses fully convolutional layers to produce spatially coherent output maps. 
Building on FCN, UNet~\citep{ronneberger2015unet} was designed primarily for biomedical image segmentation. It employed a symmetric encoder-decoder architecture with skip connections that allowed detailed spatial information from early layers to be merged with semantic information in deeper layers. UNet’s architecture has since inspired a broad range of encoder-decoder models and domain-specific adaptations, including applications to historical map segmentation~\citep{uhl_automated_2020, yuan2025semseg, wu_domain_2023}.

More recently, Vision Transformers (ViTs)~\citep{vit} have reshaped the semantic segmentation landscape by modeling long-range dependencies more effectively than CNNs. SETR~\citep{zheng2021setr} was among the first to fully replace CNN backbones with pure transformer encoders for semantic segmentation. It treated images as sequences of patches and used a transformer to model their global relationships, followed by simple upsampling modules to produce segmentation maps. Hybrid models like Swin Transformer~\citep{liu2021swin} introduced hierarchical attention mechanisms and shifted windows, achieving strong performance while maintaining scalability. SegFormer~\citep{xie2021segformer} further refined this design by combining efficient attention with lightweight decoders, enabling fast inference with competitive accuracy.
These transformer-based models have outperformed traditional CNNs in many benchmarks, especially when trained on large-scale datasets, highlighting the importance of global context and data efficiency in semantic segmentation tasks. In this paper, we fine-tune a Vision Transformer-based architecture, specifically the Segment Anything Model (SAM)~\citep{sam}, for the semantic segmentation of historical maps.


\subsection{Foundation models}
A \textit{foundation model}~\citep{bommasani_foundation_2022} is a deep learning model trained on large, diverse datasets to enable adaptation across a wide range of downstream tasks. In the context of computer vision, the introduction of the large-scale ImageNet dataset~\citep{imagenet} enabled the supervised training of foundation models such as ResNet~\citep{resnet} and vision transformers~\citep{vit, chen2021mocov3}. To learn more generalizable features, both model capacity and dataset scale have been significantly increased. Leveraging vast amounts of internet data, CLIP~\citep{clip} is trained by aligning paired text and images. In contrast, DINO~\citep{dino} and DINOv2~\citep{dinov2} adopt a self-supervised learning paradigm, with DINOv2 further scaled to larger and more diverse datasets.

While many of these models are either task-agnostic or trained using surrogate objectives, SAM~\citep{sam} is trained specifically for image segmentation using annotated data, and UMSAM~\citep{unsam} extends this to a self-supervised setting. To improve performance on domain-specific data, e.g., SAMed~\citep{samed} fine-tunes SAM on a large corpus of medical images. Building on CLIP, CLIPSeg~\citep{clipseg} introduces a segmentation model that generates output conditioned on a textual prompt specifying the target object.

Although foundation models can learn highly generalizable features and have demonstrated effectiveness across a wide range of downstream tasks~\citep{lin2024sam, zhou2024dino, luo2022clip4clip}, they often underperform in historical map segmentation. This is primarily due to the abstract visual patterns and stylistic variations in historical maps, which frequently diverge from the common-sense or conceptual priors embedded in these models. Effectively addressing this task requires a stronger integration of domain-specific prior knowledge.
To this end, MapSAM~\citep{mapsam} fine-tuned SAM on a dataset derived from the Topographic Atlas of Switzerland (Siegfried Map). The study showed that adapting a foundation model to historical maps significantly outperforms the domain-specific UNet architecture~\citep{ronneberger2015unet} when ground-truth labels are very limited. To eliminate the need for manual annotations (e.g., such as points or bounding boxes typically required by SAM), MapSAM introduced an automated module for generating prompting masks, enabling fully automatic binary segmentation. However, this approach is not directly extensible to multi-class semantic segmentation, as it requires training separate models for each class.

To address this limitation, we propose a method that fine-tunes SAM to support multi-class segmentation through prompt-based prior encoding. Specifically, we introduce a strategy that uses a single labeled example image as a prompt to encode class-specific information. This enables the model to generalize across different classes and map atlases, and to dynamically switch target classes via the prompt—facilitating more scalable and flexible semantic segmentation.

\section{Method}
\subsection{Task formalization}\label{sec:task_formal}
We focus on the semantic segmentation of topographical historical maps. These maps, originating from different collections, publishers, or time periods, often employ varying symbolic representations for the same geographic features. Our objective is to train a single model capable of segmenting a predefined target class $X$ from large map sheets (e.g., $10000\times 10000$ pixels) by providing the model with only a single example of the target class $X$: a small cropped source image $S_I$ (e.g., $384\times384$ pixels) that contains pixels of class $X$, along with its corresponding segmentation mask $S_X$. The target class $X$ is randomly selected from a class set $ \mathcal{C}=\{c_1, c_2, \dots, c_K\}$, where $K$ is the total number of classes. In this work, we train the model on two datasets Gauß and Siegfried data which contains 7 classes in total. More details are in section \ref{sec:data_main}.

\subsection{OND-knowledge-based Prompting}
Pre-trained foundation models offer promising feature representations for downstream zero- or few-shot tasks on new datasets. Through prompting, one can elicit task-relevant information from the model. Inspired by this technique, SAM~\citep{sam} introduces the concept of a \textit{promptable segmentation task}, where spatial cues—such as points or bounding boxes—are used as prompts to guide the model in generating segmentation masks for the target content.

However, such spatial-clue prompting proves insufficient for accurately extracting all relevant content in historical maps (see Figure~\ref{fig:sam_clip}). Fine-tuning SAM on historical map data may help the model segmenting the content within a small cropped region corresponding to the provided spatial clues of prompt. Yet, this approach often fails to efficiently segment all the target content in the whole map sheet, where instances of the target class may appear in varying shapes, sizes, and locations, often scattered or disconnected. For example, point-based prompting would require numerous individual prompts to approximate the spatial distribution of the target class. Similarly, bounding-box prompts struggle to distinguish foreground from background in cases where multiple patterns coexist within the same region. To overcome these limitations, we propose encoding OND knowledge as a form of prompting by explicitly informing the model what pattern to extract, in terms of  an image and its corresponding segmentation mask.

\subsection{Weight-Decomposed Low-Rank Adaptation (DoRA)}
To efficiently adapt the large-scale image encoder of SAM to the domain of historical maps, we adopt Weight-Decomposed Low-Rank Adaptation (DoRA, \cite{dora}), a parameter-efficient fine-tuning strategy that introduces minimal additional parameters while preserving model capacity. Traditional full fine-tuning is computationally expensive and memory-intensive, particularly for vision foundation models such as SAM. DoRA addresses this by decomposing the weight update into a low-rank representation and applying it in a residual manner to the original weights. Given a pre-trained weight matrix $\mathbf{W}_0\in \mathbb{R}^{d\times k}$ in the image encoder, DoRA decomposes it into a magnitude component $\mathbf{m}\in \mathbb{R}^{1\times k}$ and a direction component $\mathbf{D}\in \mathbb{R}^{d\times k}$, such that
\begin{equation}
\mathbf{W}_0 = \|\mathbf{W}_0\|_c\cdot\frac{\mathbf{W}_0}{\|\mathbf{W}_0\|_c} =\mathbf{m}\cdot\mathbf{D}, \quad \|\mathbf{D}_{:,j}\|_2 = 1 \quad \forall j,
\end{equation}
where $\|\cdot\|_c$ is the vector-wise norm of a matrix across each column. Then the weight matrix $\mathbf{W}_0$ is re-parameterized with decomposed parts $\Delta\mathbf{m}$, $\mathbf{D}$, $\mathbf{A}$ and $\mathbf{B}$, as shown in equation \ref{eq:dora_update}, where underline denotes trainable parameters.
Specifically, DoRA updates the direction via low‑rank adaptation while also allowing magnitude adjustments:
\begin{equation}\label{eq:dora_update}
\mathbf{W} = \bigl(\mathbf{m} + \underline{\Delta \mathbf{m}}\bigr)\, \bigl(\mathbf{D} + \underline{\mathbf{A}\mathbf{B}}\bigr) 
\end{equation}
where $\mathbf{A}\in \mathbb{R}^{d\times r}$ and $\mathbf{B}\in \mathbb{R}^{r\times k}$ represents the LoRA‑style~\citep{hu2022lora} low‑rank update matrices (with rank $r \ll \min\{d, k\}$), and $\Delta \mathbf{m}$ is a trainable magnitude residual. During training, only $\Delta \mathbf{m}$ and the low‑rank factors $\mathbf{A}, \mathbf{B}$ are optimized, while $\mathbf{m}$ and $\mathbf{D}$ remain frozen. This formulation preserves inference efficiency (no additional latency after merging updates) while yielding superior adaptation capacity and training stability compared to LoRA~\citep{hu2022lora}.

\subsection{Overall framework}
\begin{figure*}[t]
    \centering
    \includegraphics[width=1.0\linewidth]{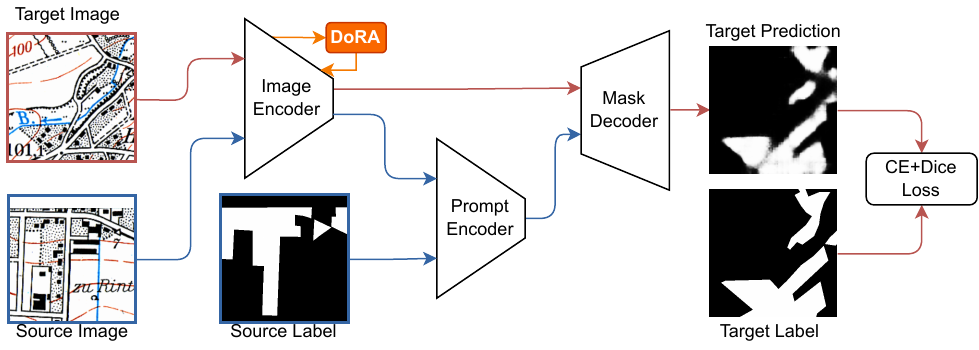}
    \caption{Overview of the proposed SMOL-MapSeg framework. Given a labeled source image and an unlabeled target image, both are processed through a shared Image Encoder (adapted through DoRA). The source image features and label are further encoded via the Prompt Encoder to capture OND knowledge. These prompt features, combined with target image features, are passed to the Mask Decoder, which predicts the segmentation mask in the target image based on the visual pattern indicated in the source label.}\label{fig:framework}
\end{figure*}

As illustrated in Figure~\ref{fig:framework}, our proposed framework, named \textit{SMOL-MapSeg}, is a modified version of SAM~\citep{sam}. Unlike the original SAM, which processes a single image, SMOL-MapSeg takes two images as input: a source image and a target image. Features from both images are first extracted using a shared \textit{Image Encoder} which is efficiently fine-tuned with DoRA~\citep{dora}. The source image features, along with its corresponding segmentation label, are then processed by a \textit{Prompt Encoder}, which encodes the OND knowledge. These prompt features are subsequently combined with the target image features and fed into the \textit{Mask Decoder}. The decoder outputs a segmentation mask for the target image, highlighting regions that share visual patterns with the labeled class in the source image.

We adopt the architecture of the \textit{Image Encoder} and \textit{Mask Decoder} from SAM~\citep{sam}, and introduce DoRA attached to the \textit{Image Encoder} for fine-tuning the encoder parameters. Then we modify the \textit{Prompt Encoder} with Equations~\ref{eq:prompt_enc_1}--\ref{eq:prompt_enc_3} to encode the source image–label pair. Let $L$ denote the binary mask corresponding to the source image. The function $f$ consists of three convolutional layers, where the first two layers are followed by Layer Normalization and GELU activation. These two layers use stride-two convolutions, resulting in a spatially $4\times$ downsampled feature map $F_L$, aligned in resolution with the source image features $F_I$. This enables element-wise multiplication ($\odot$) to fuse $F_L$ and $F_I$, as shown in Equation~\ref{eq:prompt_enc_2}. Finally, Equation~\ref{eq:prompt_enc_3} applies two additional convolutional layers to further integrate the fused features and project them into a new feature space suitable for downstream segmentation. Based on the proposed architecture, we freeze the parameters of \textit{Image Encoder}, and train the DoRA, \textit{Mask Decoder}, and \textit{Prompt Encoder} parameters of SMOL-MapSeg.

\begin{align}
    F_L = f(L) \label{eq:prompt_enc_1} \\
    F = F_I \odot F_L\label{eq:prompt_enc_2} \\
    F = g(F) \label{eq:prompt_enc_3}
\end{align}

\subsection{Sampling Source--Target Image Pairs}
To effectively train the model, source–target image pairs must be sampled with care. We define two categories of such training pairs: \textbf{positive pairs} and \textbf{negative pairs}. A positive pair consists of a randomly selected target image and a source image whose labeled foreground class is also present in the target image. In contrast, a negative pair contains a source image whose labeled class does \textit{not} appear in the corresponding target image. This pairing strategy enables the model to not only learn to identify patterns similar to those shown in the source example, but also to suppress false positives when no matching pattern is present in the target image.

\begin{algorithm}[t]
\caption{Sampling Source--Target Image Pair}
\label{alg:sampling}
\begin{algorithmic}[1]
\Require Dataset $\mathcal{D}=\{(M_1, G_1), \dots, (M_N, G_N)\}$ with $N$ map sheets and their corresponding ground-truth labels. Each map sheet $M_i$ is cropped into $N_i$ image-label pairs $(I, L)$, where $i$ is the map sheet index.
\Ensure A source--target training pair with selected class $X$ (positive or negative)

\State $(I^T_i, L^T_i) \gets$ Randomly select a target image-label pair from $\mathcal{D}$
\State $C_T \gets$ Foreground classes in $L^T_i$
\If{$|C_T| > 0 \ \And \ \texttt{Random()} < p$} \Comment{Sample a positive pair}
    \State $X \gets$ Randomly select $c \in C_T$
    \State $(I^S_i, L^S_i) \gets$ Source image-label from $(M_i, G_i)$ with label $X$
\Else \Comment{Sample a negative pair}
    \State $X \gets$ Randomly select $c \notin C_T$
    \State $(I^S_i, L^S_i) \gets$ Source image-label from $(M_i, G_i)$ with label $X$
\EndIf
\State \Return Pair $T = \left( I^T_i, L^T_i \right), S=\left( I^S_i, L^S_i \right)$ with selected class $X$
\end{algorithmic}
\end{algorithm}

The procedure for sampling a source--target image pair is detailed in Algorithm~\ref{alg:sampling}. First, a target image-label pair is selected from a randomly chosen map sheet $(M_i, G_i)$ (line 1). Next, all foreground classes present in the selected target image are identified as $C_T$ (line 2). If any class $X\in \mathcal{C}$ is found in $C_T$, a positive pair is sampled with probability $p$ (empirically set to 0.75). As shown in lines 4–5, one target class $X$ is randomly selected from $C_T$, and a corresponding source image is retrieved from the same map sheet that also contains class $X$. This ensures that both the source and target images use the same symbolization pattern to represent class $X$.
If the condition in line 3 is not satisfied (i.e., $C_T=\emptyset$ or the random sampling falls outside probability $p$), a negative pair is constructed. In this case, a class $X$ is randomly selected such that it is absent from the target image but present in the source image. This encourages the model to suppress false positives when the desired pattern is not present in the target image.

\section{Dataset}
\subsection{Main datasets}\label{sec:data_main}

The four datasets used in this study are summarized in Table~\ref{tab:dataset_summary}, comprising two collections from Germany and two from Switzerland. These maps originate from different cartographic sources and historical periods, each employing distinct figurative styles to depict topographical features. As a result, objects belonging to the same semantic class may exhibit varying visual patterns across datasets. In total, six semantic classes are annotated: WL (Woodland), GL (Grassland), SM (Settlement), WT (Water), RW (Railway), and VY (Vineyard). Each map collection includes a different subset of these classes, and the class-wise proportions among foreground pixels vary within each subset (see the last row of Table~\ref{tab:dataset_summary}). For example, in \textit{Hameln} dataset the minority WT class constitutes only 3\% of the foreground pixels. Despite these heterogeneous class configurations, the proposed OND-based prompting strategy enables training a single SMOL-MapSeg model across all datasets. For comparative analysis, the model is also trained on subsets of these datasets to evaluate performance under partial domain exposure. Note that the validation sets of \textit{Siegfried-RW} and \textit{Siegfried-VY} are not used for model selection during fine-tuning. This is because our objective is to adapt the foundation model to the target domain using the minimal amount of labeled data. Relying on additional labeled validation data for model selection would contradict this goal. Instead, we select the best-performing model based on the training loss, aligning with the principle of achieving strong performance under minimal supervision.

\begin{table}[ht]
    \centering
    \caption{Overview of datasets.}
    \begin{tabularx}{\linewidth}{lXXXX}
        \toprule
        \textbf{Dataset} & \textit{Hameln} & \textit{Donauwörth} & \textit{Siegfried-RW} & \textit{Siegfried-VY} \\
        \midrule
        \textbf{Source} 
        & Topographic Maps (TK) in Hameln, Lower Saxony, Germany 
        & Topographic Maps (TK) in Donauwörth, Bavaria, Germany 
        & Swiss Siegfried Maps with Railway annotations 
        & Swiss Siegfried Maps with Vineyard annotations \\\hline
        \textbf{Num. of tiles} 
        & 15,300 
        & 2,376 
        & 8,390 
        & 877 \\\hline
        \textbf{Tile size} 
        & $384 \times 384$ 
        & $384 \times 384$ 
        & $224 \times 224$ 
        & $224 \times 224$ \\\hline
        \textbf{Years}
        & 1897 -- 2017
        & 1910 -- 1959
        & 1870 -- 1926
        & 1870 -- 1926 \\\hline
        \textbf{Train:Val:Test}
        & 5:0:5
        & 5:0:5
        & 7:1:2
        & 7:1:2\\\hline
        \textbf{Class:Ratio}
        & WL:61\%, GL:18\%, SM:18\%, WT:3\%
        & WL:44\%, GL:53\%, WT:4\%
        & RW:100\%
        & VY:100\%\\
        \bottomrule
    \end{tabularx}
    \label{tab:dataset_summary}
\end{table} 

\subsection{Few-shot dataset}\label{sec:data_fewshot}
For the few-shot model adaptation experiment we select three map sheets from the main dataset, namely the \textit{Hameln} map sheet in the years 1898, 1974 and 2002. They span a large time period and have very different symbolic styles. We crop an image patch that contains the new target class from the original map sheet. The details of the cropped images and the labeled new classes are summarized in Table~\ref{tab:data_fewshot}. For example, in the year 1898, we crop an image patch of size $1716\times 1058$, which is only $1.4\%$ of the original map sheet. This patch contains about $27$K labeled pixels ($19.6\%$ out of the whole patch) which describe the \textit{Building} class.
Similarly, we crop a smaller patch on the map of the year 1974. For the map in the year 2002, we crop a larger patch in order to include enough of all new target classes \textit{Industrial area}, \textit{Sport area} and \textit{Federal highway} and \textit{State road}.

To prepare the data for fine-tuning, we generate images of the same size as the main dataset. Totally, we obtained 48 training and 50 test sample images as shown in the last column of Table~\ref{tab:data_fewshot}.

\begin{table}[]
    \centering
    \caption{Labels for few-shot adaptation.}\label{tab:data_fewshot}
    \begin{tabular}{lllll}
    \hline
     year &  patch size &  label & \#labeled pixels & \#train/test images\\\hline
     1898 & 1716 $\times$ 1058 (1.4\%) & Building & 27,432 (1.5\%) & 3/4\\\hline
     1974 & 801 $\times$ 558 (0.3\%) & Building & 87,717 (19.6\%) & 3/3\\\hline
      \multirow{4}{*}{2002} & \multirow{4}{*}{3718 $\times$ 2060 (6\%)}  & Industrial area & 774,163 (10.1\%) & \multirow{4}{*}{42/43} \\
      &   & Sport area & 363,876 (4.7\%)\\ 
      &  & Federal highway & 137,189 (1.8\%)\\
       &  & State road & 81,197 (1.1\%)\\
     \bottomrule
    \end{tabular}
\end{table}

\section{Experiment}\label{sec:experiment}
\subsection{Experiment settings} 
The SMOL-MapSeg model is initialized with the pre-trained parameters from the "ViT-B" variant of SAM. We then fine-tune the parameters of DoRA, the prompt encoder, and the mask encoder for 200 epochs (or 100,000 iterations if the total number of iterations is fewer than 100,000), using a batch size of 16 with gradient accumulation over 4 steps. This results in an effective batch size of $64$. This approach allows us to train the model with a larger effective batch size while operating within limited GPU resources. All map tiles are zero-padded to a fixed input size of $384 \times 384$ when smaller than this resolution. Following the original SAM training strategy, we optimize the model using a combination of binary cross-entropy (BCE) loss and Dice loss.

To evaluate the generalization capability of SMOL-MapSeg, we further fine-tune the trained model for 50,000 iterations on a few-shot dataset. We refer to this process as few-shot model adaptation of SMOL-MapSeg, as it involves only a limited number of new samples. Details of the few-shot datasets are provided in Section~\ref{sec:data_fewshot}.

\subsection{Training details} 

We use the Adam optimizer with a base learning rate of $b = 5 \times 10^{-4}$ and a weight decay of 0.01 to update the model parameters. To ensure training stability, the learning rate follows a warm-up schedule for the first 1,000 iterations, after which it is gradually decayed according to the following schedule:

\begin{equation}
    \text{lr} = 
    \begin{cases}
    \frac{t \times b}{T_w} & \text{for } t \leq T_w \\
    b \cdot \left(1 - \frac{t - T_w}{T_{\text{max}}} \right)^{0.9} & \text{for } t > T_w
    \end{cases}
\end{equation}
Here, $b$ denotes the base learning rate, $t$ is the current training iteration, $T_w$ is the number of warm-up iterations, and $T_{\text{max}}$ is the total number of training iterations. All experiments are conducted using a single NVIDIA RTX 4090 GPU.

\section{Evaluation metrics}
Since segmentation masks are generated by providing the model with a single source image–label pair as a prompt, we evaluate the segmentation performance separately for each class. We use three standard metrics: Intersection over Union (IoU), Precision, and Recall, as defined in Equations~\ref{eq:iou}--\ref{eq:rec}. Let $TP=\{\hat{y}|\hat{y}=y\}$ denote the set of true positive (foreground) pixels correctly predicted. Let $\hat{Y}_\text{pos}$ be the set of predicted foreground pixels and $Y$  the set of ground-truth foreground pixels. The notation $|\cdot|$ denotes the number of pixels in a set.

\begin{align}
    IoU = {|TP|}/{|\hat{Y}_\text{pos} \cup Y|}\label{eq:iou} \\
    Prec ={|TP|}/{|\hat{Y}_\text{pos}|}\label{eq:prec} \\
    Rec = {|TP|}/{|Y|}\label{eq:rec} 
\end{align}

\section{Results and Discussions}
\subsection{Comparison to baseline models}
Table~\ref{tab:siegfried_compare} presents the IoU comparison between SMOL-MapSeg and several baseline models, including SAMed~\citep{samed} and Few-Shot SAM~\citep{fewshot_sam}, which are designed for fine-tuning SAM on medical images, as well as MapSAM~\citep{mapsam}, which adapts SAM for binary segmentation of historical maps. All baseline models are fine-tuned from SAM using the two Siegfried datasets.
For a fair comparison, SMOL-MapSeg adopts the same configuration from MapSAM in terms of batch size, total training epochs, and DoRA rank. This configuration differs from the default settings defined in Section~\ref{sec:experiment}. Besides, the baseline results are adopted from MapSAM.
On average, SMOL-MapSeg achieves the highest performance among all models. Notably, it exhibits a substantial improvement over MapSAM, the second-best performing method.

\begin{table}[t]
\centering
\setlength{\tabcolsep}{6pt}
\resizebox{\linewidth}{!}{
\begin{tabular}{l *{6}{c}}
\toprule
\multicolumn{1}{c}{} &
\multicolumn{4}{c}{\textbf{Siegfried-RW}} &
\multicolumn{2}{c}{\textbf{Siegfried-VY}} \\
\cmidrule(lr){2-5}\cmidrule(lr){6-7}
\textbf{Model} &
Full (5872) & 10\% (587) & 1\% (58) & 10-shot &
Full (613) & 10-shot \\
\midrule
UNet
& \textbf{91.86} & \textbf{90.56} & 83.52 & 61.43
& \underline{77.04} & 60.23 \\
SAMed
& 86.31 & 85.69 & 86.01 & 75.44
& 74.85 & \underline{61.53} \\
Few-Shot SAM
& -- & -- & -- & 35.82
& -- & 46.83 \\
MapSAM
& 89.46 & 88.71 & \underline{86.53} & \underline{78.50}
& 74.32 & 60.02 \\
SMOL-MapSeg
& \underline{90.90} & \underline{90.55} & \textbf{86.59} & \textbf{82.50}
& \textbf{81.05}  & \textbf{62.29} \\
\bottomrule
\end{tabular}
}
\caption{IoU (in \%) comparison of binary classification between SMOL-MapSeg and baseline models on Siegfried dataset.}
\label{tab:siegfried_compare}
\end{table}

\begin{figure}[b]
    \centering
    \includegraphics[width=\linewidth]{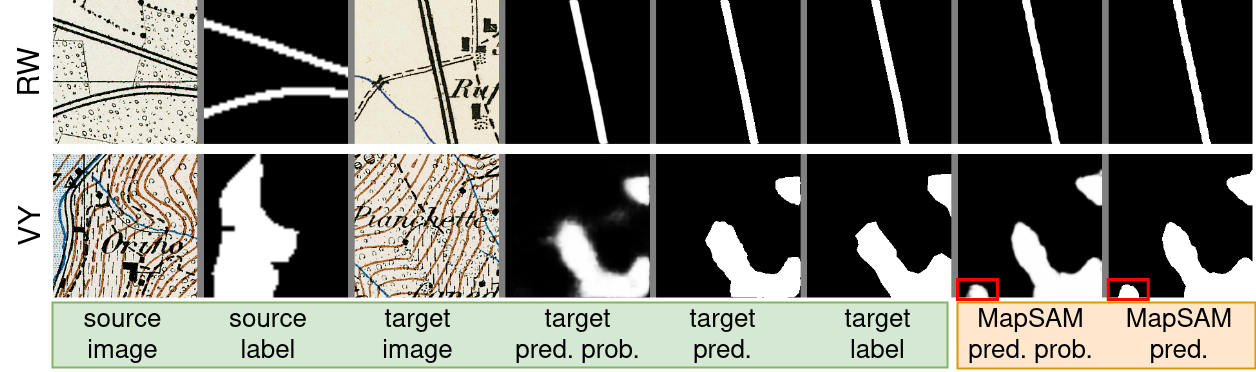}
    \caption{Segmentation results of SMOL-MapSeg (Columns with green box in the bottom) and MapSAM (Columns with orange box in the bottom) on Siegfried datasets.}
    \label{fig:smolseg_vs_mapsam}
\end{figure}

\begin{table}[t]
    \centering
    \resizebox{\linewidth}{!}{
    \begin{tabular}{*{2}{c}|*{4}{c}|*{3}{c}|c|c|c}
    \hline
    \multicolumn{2}{c|}{Dataset} & \multicolumn{4}{c|}{Hameln} & \multicolumn{3}{c|}{Donauwörth} & S-RW & S-VY & \multirow{2}{*}{Mean} \\\cline{1-11}
    \multicolumn{2}{c|}{Class} & WL & GL & SM & WT & WL &GL &WT & RW & VY\\\hline
    \multirow{3}{*}{IoU}
    &MapSAM & - & - & - & - & - & - & - & 89.46 & 74.32 & -\\
     &SMOL-Uni & \textbf{95.00} & 74.33 & 85.07 & 65.38 & 82.20 & 82.69 & 53.75 & 88.36 & 75.80 & 78.06\\
     &SMOL-Ind & 94.04 & 72.46 & 83.31 & 59.70 & 74.23 & 82.71 & 68.33 & \textbf{92.46} & \textbf{79.96} & \textbf{78.58}\\\hline
    \multirow{2}{*}{Prec.}
     &SMOL-Uni & 96.97 & 81.64 & 92.74 & 83.66 & 90.08 & 90.43 & 86.22 & 95.63 & 87.50 & \textbf{89.43}\\
     &SMOL-Ind & 96.64 & 81.65 & 89.10 & 76.95 & 84.52 & 91.70 & 88.91 & 96.32 & 87.34 & 88.13\\\hline
     \multirow{2}{*}{Rec.}
     &SMOL-Uni & 97.90 & 89.25 & 91.15 & 74.95 & 90.39 & 90.62 & 58.80 & 92.08 & 85.00 & 85.57\\
     &SMOL-Ind & 97.22 & 86.55 & 92.76 & 72.70 & 85.91 & 89.40 & 74.69 & 95.85 & 90.44 & \textbf{87.28}\\\hline
    \end{tabular}
    }
    \caption{Comparison between universal training and independent training. All metrics are shown in \%.}
    \label{tab:uni_vs_ind}
\end{table}

The qualitative comparison is shown in Figure~\ref{fig:smolseg_vs_mapsam}. For the railway (RW) class, SMOL-MapSeg and MapSAM produce comparable results. In the vineyard (VY) region, SMOL-MapSeg demonstrates stronger reliability under ambiguous boundaries: in the absence of clear linear cues in the input images, it expresses uncertainty through well-calibrated, spatially diffuse target probabilities rather than imposing overly crisp contours. This uncertainty-aware behavior curbs false positives and avoids over-segmentation, whereas MapSAM often yields overconfident, sharp boundaries that misalign with the ground truth and spuriously label some background pixels as vineyard (see red boxes in the last two columns of Fig.~\ref{fig:smolseg_vs_mapsam}).

\begin{figure}[p]
    \centering
    \begin{minipage}{\linewidth}
        \centering
        \includegraphics[width=0.7\linewidth]{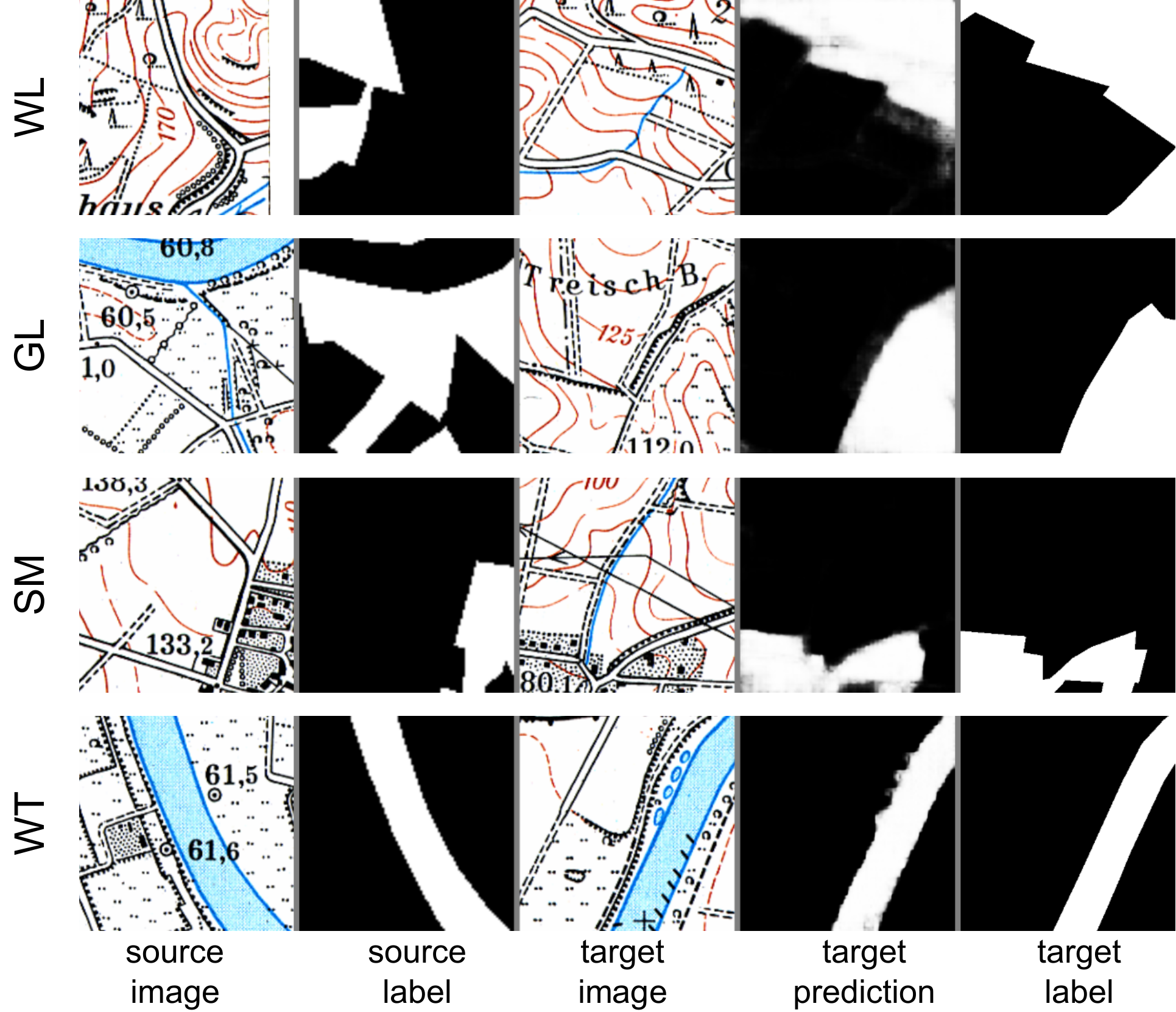}
        \caption{Segmentation results of SMOL-MapSeg on Hameln dataset.}
        \label{fig:res_hameln}
    \end{minipage}

    \vspace{0.3cm}

    \begin{minipage}{\linewidth}
        \centering
        \includegraphics[width=0.7\linewidth]{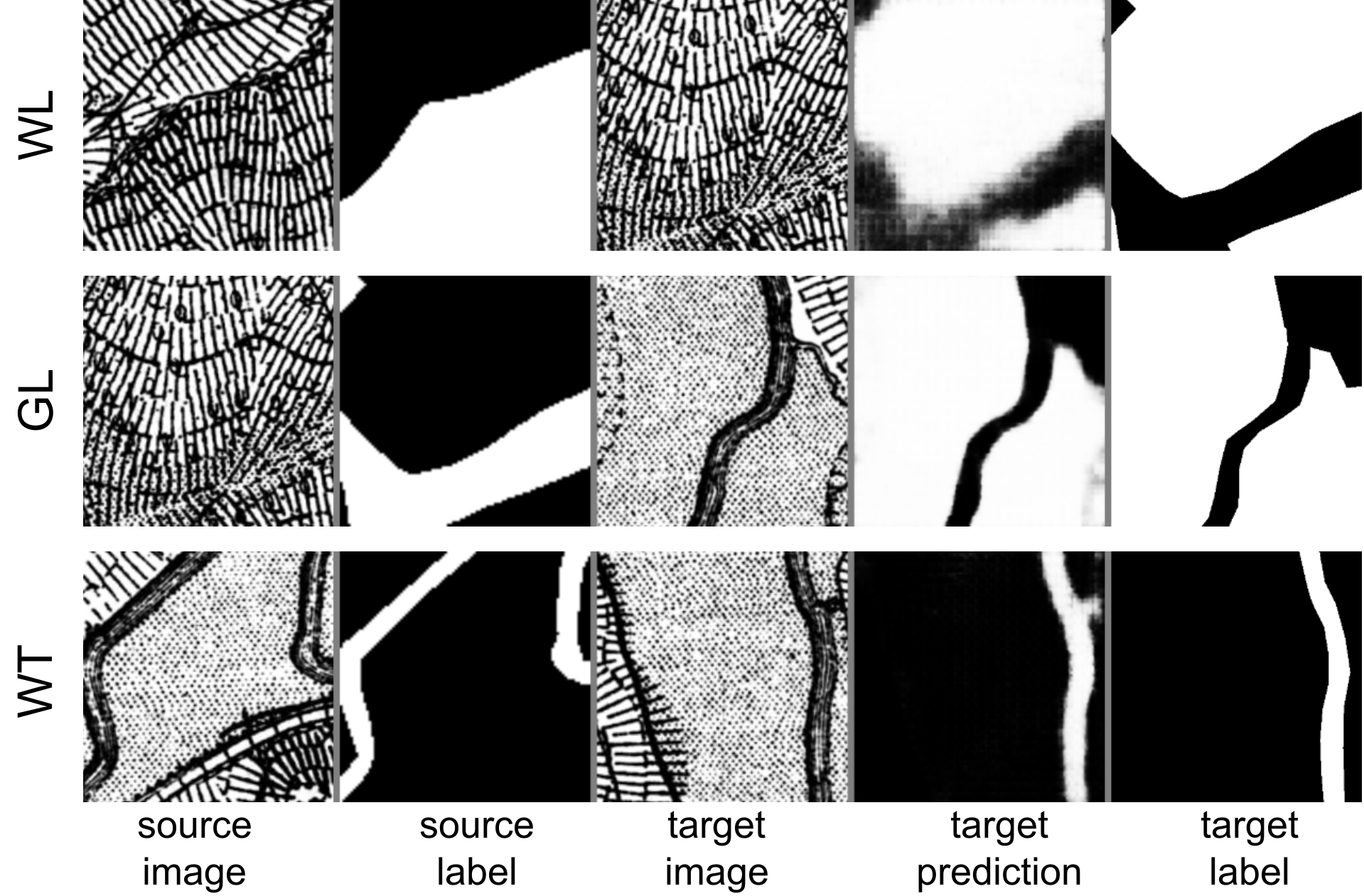}
        \caption{Segmentation results of SMOL-MapSeg on Donauwörth dataset.}
        \label{fig:res_bayern}
    \end{minipage}
\end{figure}
\subsection{Unified model for arbitrary classes and datasets}
Baseline model, MapSAM, is designed for binary segmentation and requires separate models to be trained for different datasets or classes. In contrast, our proposed SMOL-MapSeg framework is capable of segmenting any class guided by an OND-knowledge-based prompt, enabling a more flexible and unified approach. Specifically, a single SMOL-MapSeg model can be trained across arbitrary semantic segmentation datasets—each with distinct target classes.

Qualitative results of this unified model across four datasets are shown in Figures~\ref{fig:smolseg_vs_mapsam}–\ref{fig:res_bayern}. On test samples from all datasets, the model successfully segments the correct class as guided by OND knowledge, even for the Siegfried-VY dataset, which contains very limited training data. 

Table~\ref{tab:uni_vs_ind} presents quantitative comparisons of SMOL-MapSeg under two training configurations using the same model architecture and hyperparameters but different datasets. The rows labeled \textit{SMOL-Uni} correspond to a single model trained on all datasets, while the rows labeled \textit{SMOL-Ind} correspond to models trained independently on each dataset.
When trained independently, \textit{SMOL-Ind} significantly outperforms MapSAM on both the Siegfried-RW (S-RW) and Siegfried-VY (S-VY) datasets. The unified model, \textit{SMOL-Uni}, achieves competitive performance across all datasets, with mean IoU only slightly lower than the average of the individually trained \textit{SMOL-Ind} models. Notably, \textit{SMOL-Uni} still outperforms MapSAM on average across the RW and VY classes.
Across datasets with multiple classes, both \textit{SMOL-Uni} and \textit{SMOL-Ind} perform markedly better on majority classes than on minority classes (see Table~\ref{tab:dataset_summary} for class ratios). For example, in \textit{Hameln} the IoU of WL exceeds that of WT by more than 30\%.

Comparing the last four rows of Table~\ref{tab:uni_vs_ind}, \textit{SMOL-Uni} exhibits higher precision but lower recall than \textit{SMOL-Ind}. We attribute this to per-dataset specialization: independently trained \textit{SMOL-Ind} models are more prone to overfitting their respective datasets, leading to more liberal positive predictions for the dataset-specific foreground class—thereby boosting recall at the cost of precision.

\subsection{Qualitative results with different OND knowledge}
\begin{figure*}[t]
    \centering
    \includegraphics[width=0.95\linewidth]{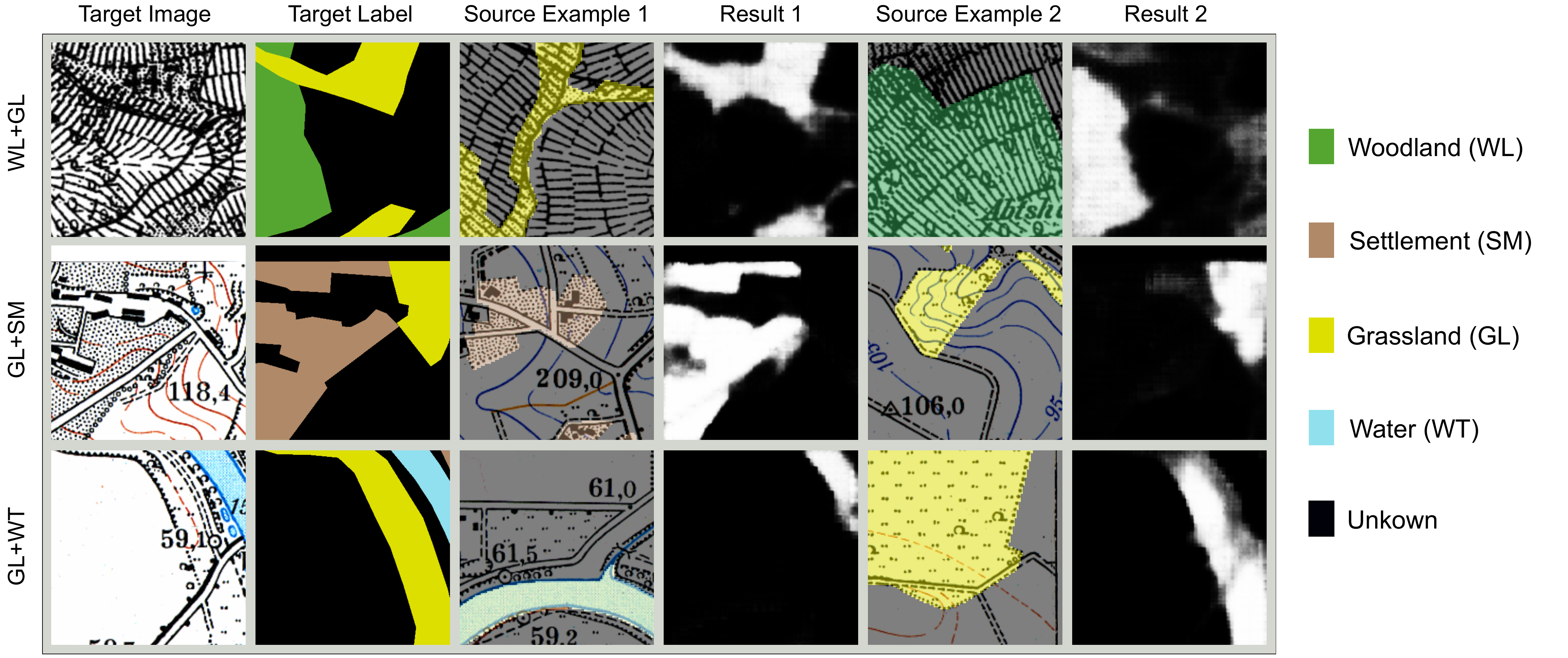}
    \caption{Segmenting the target images with source examples of different classes.}
    \label{fig:qua_2}
\end{figure*}

Figure~\ref{fig:qua_2} shows segmentation results on the same target image using different source image–label examples as OND-knowledge prompts. The selected target images contain multiple foreground classes. The model successfully demonstrates the core objective of this work: generating the corresponding segmentation mask based on the provided OND knowledge. For example, in the first row, the target image includes both the \textit{Woodland} and \textit{Grassland} classes. Although the different patterns in this image are visually subtle and difficult to distinguish at first glance (requiring zooming in to observe the details), the model successfully identifies and segments the correct patterns corresponding to the two different source examples. 

\subsection{Label efficiency}

\begin{table}[th]
    \centering
    \resizebox{\linewidth}{!}{
    \begin{tabular}{c|*{4}{c}|*{3}{c}|c|c|c}
    \hline
    Dataset & \multicolumn{4}{c|}{Hameln} & \multicolumn{3}{c|}{Donauwörth} & S-RW & S-VY & \multirow{2}{*}{Mean} \\\cline{1-10}
    Class & WL & GL & SM & WT & WL &GL &WT & RW & VY\\\hline
    10-shot (4.69\%) & 91.47 & 69.20 & 82.40 & 49.97 & 66.01 & 80.69 & 72.44 & 56.41 & 68.51 & 70.79 \\
    50-shot (20.72\%) & 93.29 & 63.75 & 81.31 & 72.26 & 74.85 & 80.81 & 64.29 & 45.90 & 73.60 & 72.23\\
    100-shot (36.17\%) & 94.27 & 64.20 & \textbf{85.82} & \textbf{76.37} & \textbf{84.06} & \textbf{84.71} & \textbf{57.07} & 81.83 & 74.32 & \textbf{78.07}\\
    all (100\%) & \textbf{95.00} & 74.33 & 85.07 & 65.38 & 82.20 & 82.69 & 53.75 & \textbf{88.36} & \textbf{75.80} & 78.06\\
    \hline
    \end{tabular}
    }
    \caption{IoU (in \%) of SMOL-MapSeg trained on different number of samples.}
    \label{tab:label_efficiency}
\end{table}

To evaluate the relationship between SMOL-MapSeg's performance and the number of training samples, we train the model using varying amounts of randomly selected data. The results are presented in Table~\ref{tab:label_efficiency}. The number of samples denoted as \textit{n-shot} refers to the minimum number of image patches selected for each class from each map sheet, ensuring that each target pattern appears at least $n$ times. For example, the \textit{Hameln} dataset comprises map sheets from different years and styles. From each map sheet (originally sized around $11\text{k} \times 11\text{k}$ pixels), $n$ cropped image samples are selected accordingly for each class. Note that each image patch may contain multiple classes, which can result in the number of selected images for some classes slightly exceeding $n$. The ratio of the final number of selected images to the total number of training images is shown in the first column of Table~\ref{tab:label_efficiency}.

\begin{figure}[t]
    \centering
    \includegraphics[width=0.8\linewidth]{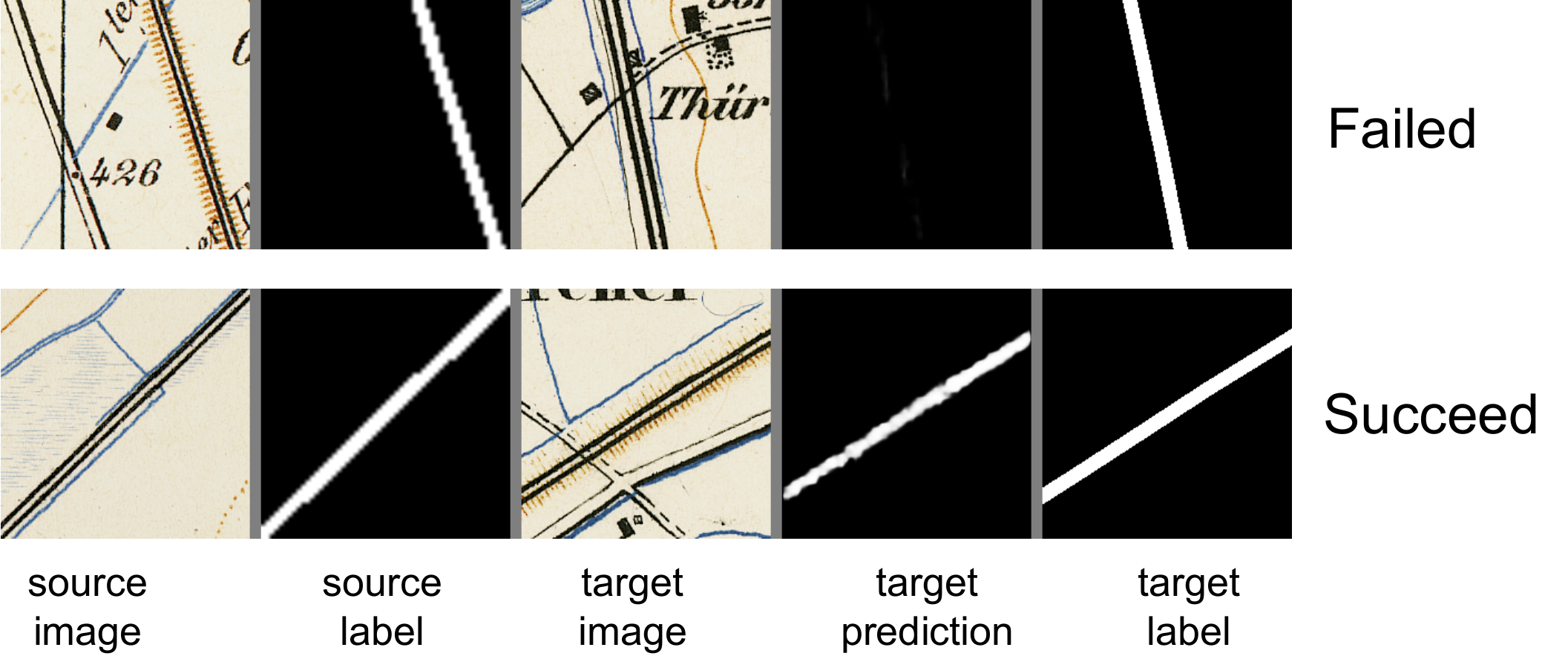}
    \caption{SMOL-MapSeg failed to segment the \textit{RW} class of the Siegfried-RW dataset.}
    \label{fig:failure_raiway}
\end{figure}

On average, the results show that training with \textit{100-shot} data, which corresponds to approximately $36.17\%$ of all available training samples, achieves performance comparable to training with the full dataset. Notably, the \textit{100-shot} model performs significantly better on minority classes, such as the \textit{WT} class in both the \textit{Hameln} and \textit{Donauwörth} datasets. However, the railway class ($RW$) shows substantially lower performance compared to the fully trained model. This reduction is likely due to the high visual similarity between the linear features of the \textit{RW} class and various background linear structures, as shown in Figure~\ref{fig:failure_raiway}. The model appears to require more training samples to effectively distinguish railway features from similar background elements. In contrast to the independently trained \textit{SMOL-Ind} model (Table~\ref{tab:uni_vs_ind}), which tends to overfit to the foreground class and therefore yields high recall, the universally trained models are more conservative in classifying ambiguous pixels as foreground, doing so only when similar patterns have been observed during training. This cautious behavior contributes to the overall lower performance for the \textit{RW} class when the number of training samples for this class is very limited.

\subsection{Few-shot model on new OND knowledge}
\begin{figure}[t]
    \centering
    \includegraphics[width=0.8\linewidth]{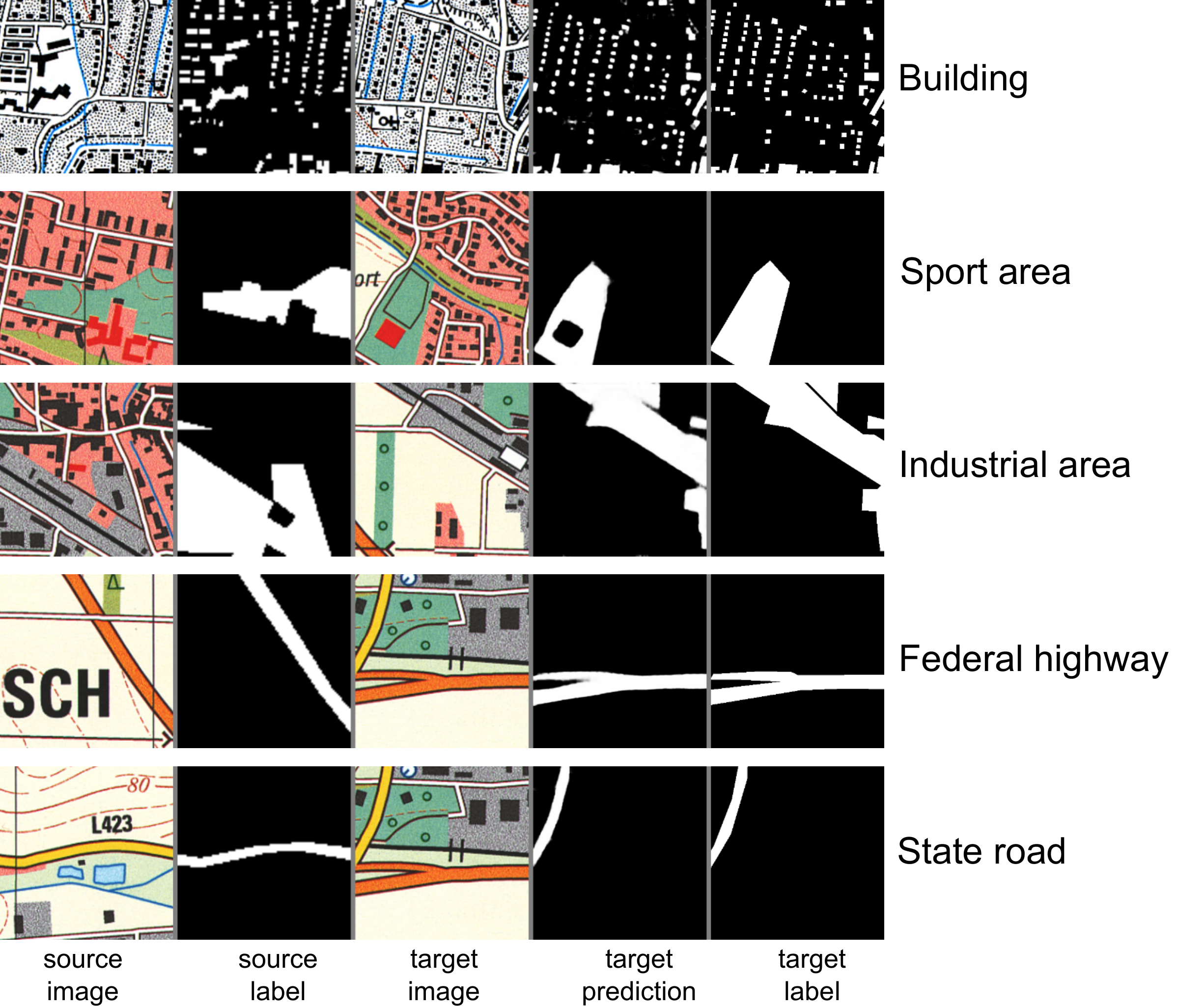}
    \caption{Few-shot adaptation on new classes.}
    \label{fig:fewshot_new}
\end{figure}

\begin{table}[]
    \centering
    \begin{tabular}{l|c|c|c}
    \hline
         &  IoU & Prec. & Recall\\\hline
        Building & 75.37 & 86.73 & 85.19\\
        Sport area & 94.87 & 97.77 & 96.97\\
        Industrial area & 97.04 & 98.71 & 98.28\\
        Federal highway & 94.23 & 96.75 & 97.32\\
        State road & 92.08 & 95.04 & 96.73\\\hline
    \end{tabular}
    \caption{Performance of fine-tuned model on few-shot new classes. All metrics are shown in \%.}
    \label{tab:fewshot_new}
\end{table}

The few-shot adaptation results on new classes are presented in Figure~\ref{fig:fewshot_new}. The first row illustrates the segmentation results for the \textit{Building} class. Although this class primarily features black and white patterns, the model is able to classify it with reasonably high accuracy through few-shot learning, achieving an IoU of $75.37\%$ as shown in Table~\ref{tab:fewshot_new}. However, due to the low-pass filtering characteristics of the image encoder, the model struggles to accurately capture the fine-grained boundaries of small building structures. In contrast, the segmentation maps in the last four rows of Figure~\ref{fig:fewshot_new} present new target classes, including \textit{Industrial area}, \textit{Sport area}, \textit{Federal highway}, and \textit{State road}, using distinct colors that make them visually more distinguishable and easier to segment. As a result, all of these classes achieve IoU scores above $90\%$.

\subsection{Failure cases}
\begin{figure}[t]
    \centering
    \includegraphics[width=0.8\linewidth]{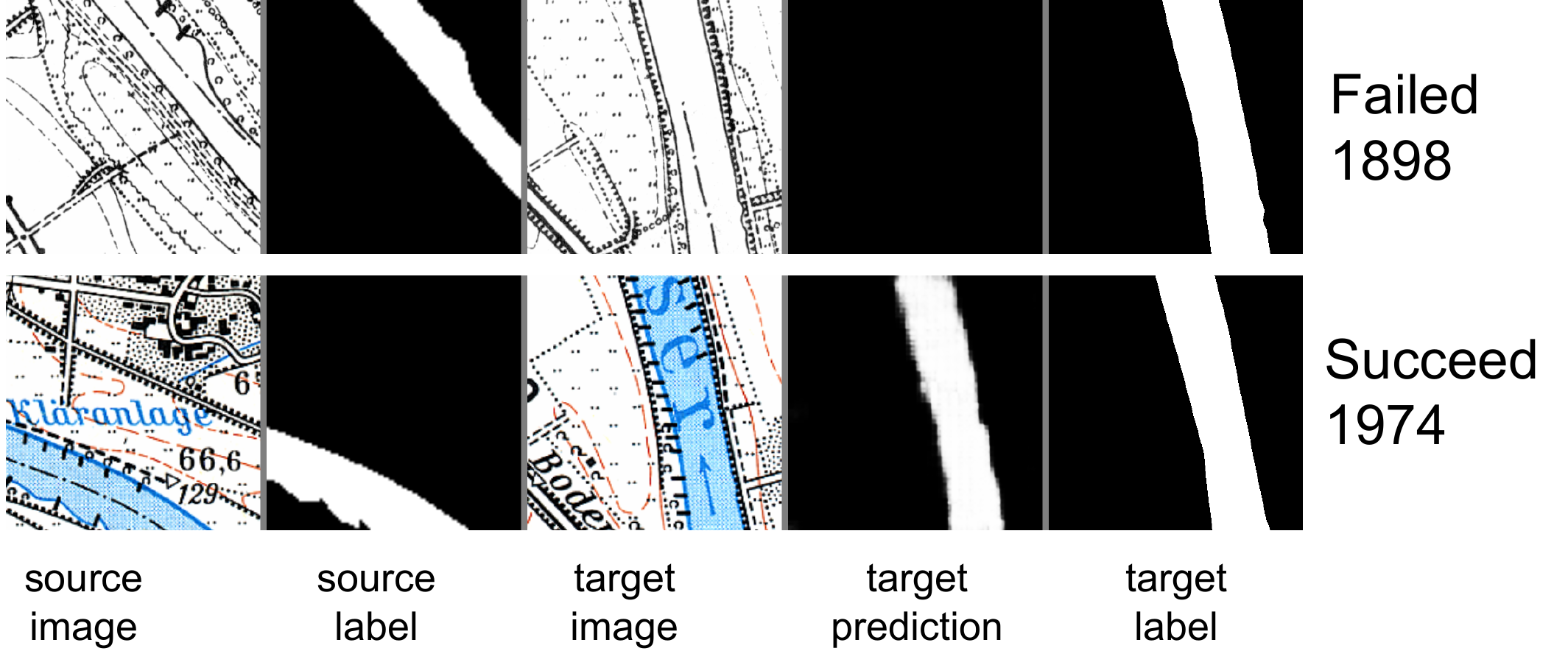}
    \caption{SMOL-MapSeg failed to segment the \textit{WT} class in the 1898 map sheet.}
    \label{fig:failure_water}
\end{figure}

In this paper, we identified two scenarios in which the model fails to correctly detect the all pixels of the foreground class. The first case, shown in Figure~\ref{fig:failure_raiway}, involves the \textit{Railway} class. The universally trained model is unable to detect all railway features due to their high visual similarity to background linear structures, as well as the diverse environments in which railways are represented. For instance, in Figure~\ref{fig:failure_raiway} (first row), railway elements overlap with orange strokes and blue lines, making them more challenging to distinguish.

The second case is illustrated in Figure~\ref{fig:failure_water}. The model is trained using only high-resolution imagery scanned from the original map sheets. These inputs are limited to image patches of size $384 \times 384$ or $224 \times 224$ pixels, which offer rich local detail but lack global spatial context. This restricted view can lead to segmentation errors, especially for classes with weak or ambiguous visual cues. For example, as shown in Figure~\ref{fig:failure}, the \textit{Water} class is not consistently identified. The same target image patch is taken from the years 1898 and 1974. In the 1974 map, the \textit{Water} class is correctly segmented due to its distinct blue coloration. However, in the 1898 version, it is rendered in white, which is visually indistinguishable from the map background. Without access to global context, the model fails to differentiate this class, leading to misclassification.

\subsection{Ablation study}
In this paper, we introduced a novel OND-knowledge-based prompt mechanism for flexible classification of arbitrary classes. Removing this prompt encoding module would eliminate the model's ability to perform such flexible classification. Therefore, instead of ablating the newly introduced prompt encoder, which is essential for achieving this functionality, we conduct an ablation study on key hyperparameters, including the number of accumulative batches used for gradient computation ($acc$) and the rank ($r$) of the DoRA adaptation matrices. The results are presented in Table~\ref{tab:ablation}.

We first fix the DoRA rank to $r=16$ and vary $acc$. The best performance is achieved with $acc=4$, corresponding to a total effective batch size of $16 \times 4 = 64$ for gradient calculation. To explore the effect of the DoRA rank, we fix $acc=4$ and evaluate the model with $r=8$, $16$, and $32$. The results show that the model performs best with a rank of $16$.

\begin{table}[th]
    \centering
    \resizebox{\linewidth}{!}{
    \begin{tabular}{*{2}{c}|*{4}{c}|*{3}{c}|c|c|c}
    \hline
    \multicolumn{2}{c|}{Dataset} & \multicolumn{4}{c|}{Hameln} & \multicolumn{3}{c|}{Donauwörth} & S-RW & S-VY & \multirow{2}{*}{mIoU}\\\cline{1-11}
    $acc$ & $r$ & WL & GL & SM & WT & WL &GL &WT & RW & VY\\\hline
    1 & 16 & \textbf{95.35} & \textbf{75.66} & 83.97 & \textbf{66.96} & 84.67 & 77.88 & 37.19 & 87.44 & 72.24 & 75.71\\
    4 & 16 & 95.00 & 74.33 & 85.07 & 65.38 & 82.20 & \textbf{82.69} & 53.75 & 88.36 & \textbf{75.80} & \textbf{78.06}\\
    8 & 16 & 94.83 & 67.32 & 82.69 & 61.16 & 79.90 & 81.50 & 53.79 & 83.63 & 74.99 & 75.53\\
    4 & 8 & 94.68 & 72.00 & 84.57 & 44.31 & 80.19 & 80.56 & \textbf{56.95} & 88.29 & 72.78 & 74.93\\
    4 & 32 & 95.16 & 73.86 & \textbf{86.35} & 46.63 & \textbf{85.95} & 81.39 & 44.64 & \textbf{88.86} & 73.76 & 75.18\\
    \hline
    \end{tabular}
    }
    \caption{Ablation study on accumulated number of batches and DoRA rank.}
    \label{tab:ablation}
\end{table}

\section{Conclusion}

In this paper, we introduced \textbf{SMOL-MapSeg}, a modified version of the segmentation foundation model SAM~\citep{sam}, enhanced with a novel prompting strategy termed \textbf{OND-knowledge-based prompting}. The On-Need Declarative (OND) knowledge provides the model with explicit guidance about what a target class is or looks like, in the form of an image-label pair. This approach is specifically tailored for segmenting historical maps, where visual representations of objects vary significantly across different map collections. Rather than relying solely on prior knowledge learned during training, SMOL-MapSeg segments objects based on declarative knowledge supplied at inference time via prompting.

Experimental results demonstrate that SMOL-MapSeg can accurately segment classes specified by OND knowledge prompts. Compared to baseline models, SMOL-MapSeg achieves significantly better performance on the Siegfried benchmark. Thanks to the flexible OND-knowledge-based prompting mechanism, SMOL-MapSeg can be applied to any dataset using a single unified model. This represents a substantial departure from previous SAM-based fine-tuning approaches, which either follow SAM to perform class-agnostic segmentation or are modified to segment a fixed set of predefined classes. In contrast, SMOL-MapSeg enables class-aware segmentation across arbitrary classes and datasets. The experimental results further show that this single model achieves performance on par with independently trained models that are specialized for each dataset. Moreover, the label efficiency experiments indicate that SMOL-MapSeg can match the performance of full-data training using only about one third of the training data.

After fine-tuning the trained SMOL-MapSeg on a small few-shot dataset, SMOL-MapSeg is able to correctly classify new classes that were not present in the main training dataset. This indicates the model’s strong potential for generalizing to new semantic segmentation tasks with limited amount of labeled data. 

Our evaluation also revealed a limitation: SMOL-MapSeg struggles to segment classes that lack distinctive features in local context. For example, in some historical maps, water bodies represented in white are visually indistinguishable from the background; the railway linear features are similar to other linear structures in the background. To address this, future work could incorporate multi-resolution map images to capture global context. Recognizing water surfaces as part of a river, for instance, may rely more on shape and spatial continuity across a larger area than on local texture or color. Additionally, multi-resolution training could serve as an effective form of data augmentation and further enhance the model’s generalization capabilities.

\section*{Data Availability}
The Hameln and Donauwörth datasets (\cite{smol_dataset}) are available in Zenodo at \url{https://doi.org/10.5281/zenodo.17340617}. The Siegfried-RW and Siegfried-VY datasets (\cite{mapsam_dataset,mapsam}) are available in ETH Research Collection at
\url{https://doi.org/10.3929/ethz-b-000691430}. 




\section*{Use of AI}
Generative AI was used to assist with language refinement. The authors take full responsibility for the content of this manuscript.

\bibliographystyle{tfv}
\bibliography{ref}
\end{document}